\pdfoutput=1

\documentclass[11pt]{article}
\usepackage{algorithm}
\usepackage{algorithmic}
\usepackage{booktabs}
\usepackage{makecell}
\usepackage{amssymb}
\usepackage{tikz}
\usepackage{pgfplots}
\usepackage{tcolorbox}  
\tcbuselibrary{breakable}
\usepackage{booktabs}
\usepackage{makecell} 
\tcbset{fontupper=\small}
\usepackage[final]{acl}
\usepackage{enumitem}
\usepackage{times}
\usepackage{latexsym}

\usepackage[T1]{fontenc}

\usepackage[utf8]{inputenc}

\usepackage{microtype}

\usepackage{inconsolata}
\usepackage{titlesec}

\usepackage{graphicx}
\usepackage{amssymb}
\usepackage{multirow}
\usepackage{algorithm}
\usepackage{algorithmic}
\usepackage{amsmath}
\usepackage{booktabs}
\usepackage{xcolor}
\usepackage{subfigure}
\usepackage[table]{xcolor}
\definecolor{em}{gray}{0.9}
\newcommand{\cem}{\cellcolor{em}}

\title{AdapTime: Enabling Adaptive Temporal Reasoning \\in Large Language Models}

\author{
    Yimin Deng\textsuperscript{1,2}\thanks{Work was conducted at Tencent Jarvis Lab.},
    Yejing Wang\textsuperscript{2},
    Zhenxi Lin\textsuperscript{3},
    \textbf{Zichuan Fu}\textsuperscript{2},
    Guoshuai Zhao\textsuperscript{1}$^{\dagger}$,
     \textbf{Derong Xu}\textsuperscript{2},
    \\
     \textbf{Yefeng Zheng\textsuperscript{4},}
     \textbf{Xiangyu Zhao\textsuperscript{2}$^{\dagger}$,}
     \textbf{Xian Wu\textsuperscript{3}$^{\dagger}$,}
     \textbf{Li Zhu\textsuperscript{1}$^{\dagger}$,}
     \textbf{Xueming Qian}\textsuperscript{1}
    \\
    \textsuperscript{1}Xi'an Jiaotong University, \textsuperscript{2}City University of Hong Kong,\\
    \textsuperscript{3}Tencent Jarvis Lab,
    \textsuperscript{4} Westlake University
    \\
  \small{
  \texttt{
  \href{mailto:dymanne@stu.xjtu.edu.cn}{dymanne@stu.xjtu.edu.cn},
    \href{mailto:guoshuai.zhao@xjtu.edu.cn}{guoshuai.zhao@xjtu.edu.cn},
     \href{mailto:xianzhao@cityu.edu.hk}{xianzhao@cityu.edu.hk}}}
     \\
     \small{
      \texttt{
    \href{mailto:kevinxwu@tencent.com}{kevinxwu@tencent.com},
    \href{mailto:zhuli@xjtu.edu.cn}{zhuli@xjtu.edu.cn}
  }}
  } 

\begin{document}
\maketitle
\begingroup
\renewcommand\thefootnote{\relax}
\footnotetext{$^{\dagger}$  Corresponding authors.}
\endgroup
\begin{abstract}
Large language models have demonstrated strong reasoning capabilities in general knowledge question answering. However, their ability to handle temporal information remains limited. 
To address this limitation, existing approaches often involve external tools or manual verification and are tailored to specific scenarios, leading to poor generalizability. 
Moreover, these methods apply a fixed pipeline to all questions, overlooking the fact that different types of temporal questions require distinct reasoning strategies, which leads to unnecessary processing for simple cases and inadequate reasoning for complex ones.
To this end, we propose AdapTime, an adaptive temporal reasoning method that dynamically executes reasoning steps based on the input context. 
Specifically, it involves three temporal reasoning actions: reformulate, rewrite and review, with an LLM planner guiding the reasoning process.
AdapTime integrates seamlessly with state-of-the-art LLMs and significantly enhances their temporal reasoning capabilities without relying on external support. Extensive experiments demonstrate the effectiveness of our approach. The code is available at \url{https://github.com/Applied-Machine-Learning-Lab/ACL2026-AdapTime}.
\end{abstract}

\section{Introduction}

\begin{figure}[t]
    \centering
    \centering
   \includegraphics[height=0.45\textwidth]{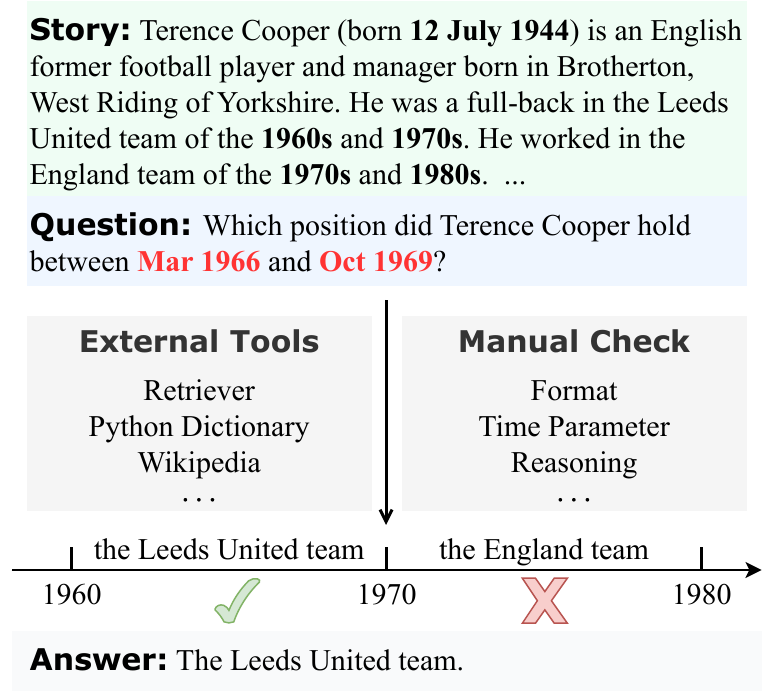}
    \caption{An example of temporal reasoning in question answering.
    }
    \label{setting}
    \vspace{-0.5cm}
\end{figure}

Recent years, large language models (LLMs) have demonstrated remarkable reasoning capabilities in general question answering~(QA) tasks~\cite{shao2023prompting,kamalloo2023evaluating,dong2024cost,bridging2025,zhang2026evoking}. However, they still face significant challenges in handling temporal questions~\cite{son2023time,xiong2024large,mesh}. Temporal reasoning tasks are concerned with understanding documents containing temporal information and answering time-sensitive questions. As shown in Figure~\ref{setting}, answering a question such as “\textit{Which position did Terence Cooper hold between Mar 1966 and Oct 1969?}” requires both retrieving relevant facts and aligning them with the temporal context. In such cases, LLMs often generate incorrect answers due to a poor understanding of temporal expressions (e.g., “between Mar 1966 and Oct 1969”)  or difficulty in identifying the order of events~\cite{chu2023timebench}. Therefore, enhancing the temporal reasoning capabilities of LLMs remains a critical challenge.

 Recent approaches typically focus on developing reasoning pipelines involve analyzing temporal questions, extracting timelines, and verifying generated answers~\cite{xiong2024large,bazaga2025learning}.
While these reasoning strategies enhance the model's ability to handle temporal information, they often rely on external support. As shown in Figure~\ref{setting} and Table \ref{tab:novelty}, one kind of approach makes use of external tools, such as applying retrievers to obtain compressed input, predefined Python dictionaries to filter time-sensitive expressions, or search engines to retrieve knowledge from sources like Wikipedia. For example,  QAaP~\cite{zhu2023question} converts questions into code and selects the optimal answer from a set of candidate entities via code execution, relying on a execution engine and manually constructed Check and Match functions. Step-back~\cite{zhengtake} combines retrieval-augmented generation with step-back prompting, requiring a retriever to obtain relevant facts. Event-AL~\cite{wu2024event} constructs a task-specific execution function and employs a Python-based solver to identify the answer. 
The other kind of approach involves manual correction during the reasoning process, including constructing timelines, verifying temporal parameters, or defining reasoning trajectories.~Time-CoT~\cite{yin2024time} addresses temporal reasoning by explicitly identifying implicit temporal parameters, organizing standard question entities along a timeline, with manual supplementation of timeline information. TG-LLM~\cite{xiong2024large} transforms textual inputs into temporal graphs (TGs) through a pipeline that includes the manual extraction of hard examples, followed by reasoning over the TGs.~TISER~\cite{bazaga2025learning} utilizes these TGs instead of raw temporal documents as context.
Although these designs can enhance temporal reasoning performance, they rely on external components that are often tailored to specific settings. Such reliance hinders their adaptability to diverse scenarios.
To address this issue, we abstract temporal reasoning into three core actions~(reformulate, rewrite, and review) that 
LLMs can autonomously perform.

\begin{table}[t]
\centering
\small
\setlength\tabcolsep{2.pt}  
\resizebox{1\linewidth}{!}{
\begin{tabular}{l|c|c|c|c|c}
\toprule
                & Reformulate   & Rewrite   & Review& Manual& External Tool    \\ \midrule
QAaP  &  \texttimes&	\checkmark&\checkmark&\checkmark&	\checkmark
 \\
    Time-CoT    &\checkmark	&\checkmark&\texttimes&\checkmark	&\texttimes

 \\ 

Event-AL        & \texttimes	&\checkmark&\checkmark&\texttimes	&\checkmark

 \\
 TG-LLM  &\texttimes	&\checkmark&\texttimes&\checkmark	&\texttimes\\
 TISER  & \texttimes	&\checkmark&\checkmark&\checkmark	&\texttimes\\
 \midrule
 AdapTime &\checkmark	&\checkmark&\checkmark&\texttimes	&\texttimes\\
\bottomrule
\end{tabular}}
\caption{The differences between our AdapTime and prior works.}
\label{tab:novelty}
\vspace{-0.5cm}
\end{table}
Moreover, different types of temporal questions may require different reasoning steps.~Simple temporal questions typically locate the answer directly from the text, while complex questions often require reasoning over multiple events and their sequential order, demanding more elaborate inference. 
To accommodate the varying levels of reasoning complexity, each of the aforementioned actions should be applied in different scenarios.
Therefore, enabling LLMs to select the appropriate action and adapt the internal reasoning strategy within that action according to the scenario and question is a key challenge in temporal reasoning.
To address this problem, we propose AdapTime, an adaptive temporal reasoning framework that dynamically selects and executes reasoning steps based on the input context and task requirements. AdapTime begins with the raw document/question and follows a multi-stage reasoning pipeline consisting of \textit{reformulate}, \textit{rewrite}, and \textit{review}. Rather than executing all steps in sequence, the model leverages the planning ability of LLMs to dynamically determine which steps to perform. This process is guided by the semantic characteristics of the question, the structural features of the input context, and the model’s confidence in its intermediate reasoning. AdapTime adapts its reasoning strategy to each instance, enabling more  accurate temporal understanding across diverse question types.

To summarize, our contributions are as follows:
\begin{itemize}[leftmargin=*,nolistsep]
\item We propose AdapTime, a novel approach for adaptive temporal reasoning, which supports a multi-stage reasoning process consisting of \textit{reformulate}, \textit{rewrite}, and \textit{review}.
\item We enable LLMs to autonomously plan and control the reasoning process, without relying on handcrafted rules, annotations, or external tools. 
\item We validate the effectiveness of our approach through comprehensive experiments on two benchmark datasets under four temporal settings.
\end{itemize}

\section{Related Work}
\subsection{LLM Reasoning}
    Large language models have made significant progress in reasoning with the ability to adapt to various downstream tasks~\cite{wang2023can,laban2023summedits,qiu2024snapntell,navigate2025,large2025}. They can generalize to new tasks through few-shot in-context learning~\cite{li2023few,dong2024survey}. Recent advances in chain-of-thought prompt~\cite{wei2022chain} strategies further demonstrate its potential in handling complex reasoning~\cite{llm4rank2025,unified2025}. By integrating multiple intermediate steps, existing methods effectively optimize the answer generation process and achieve improved reasoning performance~\cite{yeo2025demystifying,plate2023,llmemb2025,rethinking2025}. The inherent reasoning capability of LLMs has also been leveraged to enhance QA frameworks~\cite{li2024flexkbqa,stechly2024chain,mill2024a}. However, previous methods often overlook the dynamic evolution of knowledge over time, and addressing temporal questions remains for further exploration.

\subsection{Temporal Reasoning}
In recent years, a series of approaches have been proposed that employ time-sensitive reasoning pipelines to answer temporal questions. 
QAaP~\cite{zhu2023question} represents the question as a Python dictionary with predefined keys: subject, relation, object, and time. It then extracts relevant events from the document and represents them using the same format, involving executing Python code and manually defining functions.~Time-CoT~\cite{yin2024time} extracts standard question entities without temporal information
(e.g., `` Who was president of the US? '') from the question, and then searches the document for temporal parameters that match these standard entities. When the temporal parameters are implicit, they are manually matched.~TG-LLM~\cite{xiong2024large} represents text as temporal graphs~(TGs) in the form of ``(John Thompson was born in Weston) starts at 1921", consisting of a subject, relation, object, start/end, and timestamp. As LLMs may fail to produce accurate TGs, manual correction is required for some samples.~Event-AL~\cite{wu2024event} extracts relation-centric temporal events from text in the form of $(s, r, o, t)$. It also requires invoking an external Python interpreter.
These methods involve one or two of the three actions we define (reformulate, rewrite, and review), as shown in Table~\ref{tab:novelty}, and have improved the model’s temporal reasoning capabilities. However, the fixed reasoning pipeline and the reliance on handcrafted rules, external tools, or manual check restrict the model’s generalizability and adaptability across diverse temporal questions.
\section{Methods}
In this section, we first present the problem definition of temporal reasoning, then introduce each component of our proposed method, and conclude with the integration of these components into a complete reasoning process.
    \begin{figure*}[t]
    \centering
    \resizebox{1\linewidth}{!}{
    \includegraphics[height=0.7\textwidth]{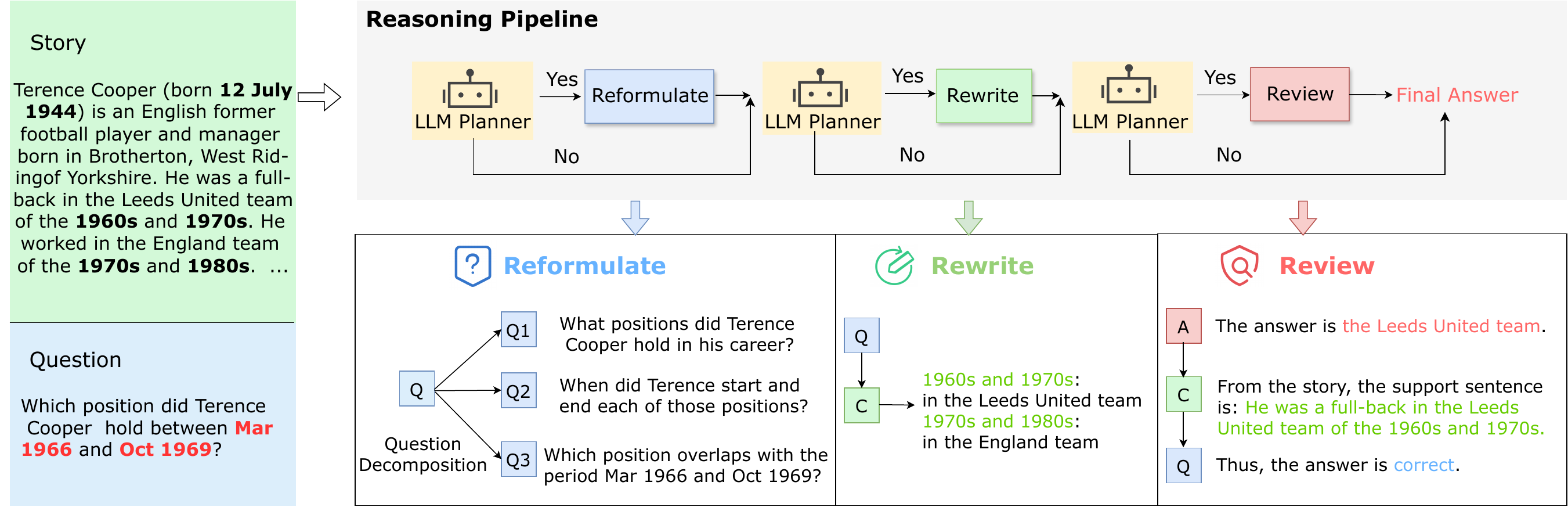}}
    \caption{The overall architecture of our model. 
    }
    \label{model}
    \vspace{-3mm}
\end{figure*}
\subsection{Problem Formulation}
The temporal question answering (TQA) task aims to answer a time-constrained question $Q$ with a specific time expression. The dataset is composed of tuples in the form $\{C, Q, A\}$, where $C$ denotes a document containing a series of temporal facts, $Q$ represents temporal questions and $A$ is the corresponding answer. The time expression in $Q$ can be either explicit, such as in the question “Who was the president of the U.S. in \textbf{1971}?”, or implicit, as in “Who was the U.S. president \textbf{before} Trump?”. 

\subsection{Overall Framework}
In this section, we provide a brief overview of our framework, which performs temporal reasoning through a flexible combination of three core operations: reformulate, rewrite, and review, as illustrated in Figure \ref{model}.
The model begins with the original document and the temporal question, and under the guidance of an LLM planner, it adaptively executes appropriate reasoning operations. First, based on the formulation of the question, the planner decides whether to decompose it into simple sub-questions. Then, if the temporal information in the context is not clear, the model transforms the relevant parts of the document into a structured format. Finally, when the model has low confidence in its answer, it extracts supporting sentences from the original document and verifies the answer. The final answer is generated after this adaptive multi-stage reasoning process.

\subsection{Reasoning Action}
In this section, we provide a detailed introduction to the three proposed actions: \textit{reformulate}, \textit{rewrite}, and \textit{review}, and explain how each contributes to the reasoning process.
\subsubsection{Reformulate}
Temporal questions often involve complex time expression and multi-hop reasoning across events, making them difficult to answer directly. Therefore, we introduce a \textit{reformulate} module that decomposes the question into a set of sub-questions, enabling more precise and interpretable reasoning.

We leverage the inherent reasoning capabilities of LLMs to perform question decomposition in a prompt-based manner. 
Formally, given an input question $Q$ and a context document $C$, the Reformulate module prompts the LLM to produce a set of sub-questions $Q = \{q_1, q_2, ..., q_n\}$, such that the answer to $Q$ can be aggregated from the answers to each $q_i\in Q$. This decomposition is adaptive and varies according to the type of the question.

For instance, as shown in Figure \ref{model}, the model handles a question involving a temporal constraint over a person’s career history. Instead of answering directly, it first extracts relevant events (positions held), then estimates their temporal spans, and finally infers the answer based on the specified time expression. This decomposition is induced by the model’s own understanding of temporal semantics, requiring no additional rule definitions or tools and thus allowing it to adapt to different settings.

The Reformulate module reduces ambiguity by isolating temporally relevant variables, allowing the model to better focus on the essential elements of the question. It also enables modular interaction with downstream components, such as retrieval and verification, which benefit from working with more focused and interpretable reasoning units. In our ablation studies, we observe that this module significantly boosts performance, especially on multi-hop and implicit temporal questions.


\subsubsection{Rewrite}
Natural language texts are usually diverse in format and often express temporal information implicitly through tense, adverbials, or discourse structure. Such implicitness makes it challenging to accurately identify event orders, durations, or overlaps.

To facilitate more accurate and consistent temporal reasoning, we introduce a \textit{rewrite} module. By rewriting the text into an explicit temporal format, such as code, timeline, or temporal graphs, we make the temporal structure more accessible for reasoning. 
This transformation is performed by an LLM prompted to reorganize and rephrase the input into a time-anchored structure. For example, implicit expressions like “during his presidency” or “after the war ended” are rewritten into explicit temporal references grounded in calendar dates or relative sequences.
 This format allows downstream modules to directly operate on time-anchored information. By aligning events with specific time points, the model can better capture their temporal order, which supports more accurate reasoning over temporal constraints.

In essence, the Rewrite module enables the system to handle complex temporal contexts with greater clarity and consistency.
By making temporal relations explicit, the Rewrite module provides a normalized and interpretable representation of temporal information. This clarity supports downstream modules such as reasoning and answering components, enabling them to carry out temporal inference with greater effectiveness and robustness.
Empirically, we find that this module significantly improves performance on questions requiring fine-grained temporal understanding, particularly in scenarios with multiple overlapping events.

\subsubsection{Review}
When generating time-sensitive answers, the model may incorrectly locate the answer within the context, especially in the presence of ambiguous or overlapping temporal information. Such mistakes can lead to factual inconsistencies, hallucinations, or misaligned reasoning steps. Therefore, verifying the generated answer is a crucial step in ensuring accuracy and consistency~\cite{stepwise2025}.

To address this, we introduce a \textit{review} module that performs a thorough validation of the model's answer by prompting it to:
\begin{itemize}[leftmargin=*,nolistsep]
\item Retrieve supporting statements from the document that justify the predicted answer.

\item Evaluate whether the supporting information is consistent with the question.

\item Revise the answer if conflicts are detected.

This module is particularly useful in implicit temporal reasoning, where the answer depends on correctly understanding multiple time-sensitive facts. By explicitly asking the model to justify its output, we introduce a review mechanism that improves robustness and interpretability.
\end{itemize}

\subsection{LLM Planner}
Temporal questions differ significantly in their complexity and the type of reasoning they require. 
Using a fixed, uniform reasoning strategy for all questions is inadequate. To accommodate this variability, we introduce a Planner module that dynamically selects the appropriate reasoning strategy based on each question and context.

The Planner is implemented as an LLM prompted to perform step-by-step decision-making. Given the reasoning context $C$ and the question $Q$, it first analyzes the temporal structure and reasoning demand, then decides which reasoning action is most suitable. Specifically, it decides whether to apply \textit{reformulate}, when the question involves multiple implicit steps that can be broken down into simpler sub-questions; \textit{rewrite}, when the temporal information in the document is implicit or scattered and needs to be transformed into a structured form; or \textit{review}, when the model lacks confidence in its answer and needs to verify it. This decision process is carried out in natural language, where the model is instructed to justify its choice and, if applicable, generate the corresponding sub-steps for execution.

By explicitly planning the reasoning process, this module allows the system to adapt to the nature of each question and enhances reasoning ability by allowing the model to focus on relevant temporal operations. 
Overall, the Planner acts as a controller that orchestrates different reasoning actions, leading to robust temporal understanding.

\begin{algorithm}[ht]
\caption{Temporal QA with Adaptive Reasoning}
\label{alg:temporal_qa_global_rewrite}
\begin{algorithmic}[1]
\REQUIRE Temporal question $Q$, document context $C$
\ENSURE Final answer $A$

\STATE $A \leftarrow \texttt{null}$

\STATE \textbf{// Step 1: Reformulate }
\STATE $d_1 \leftarrow \text{Planner}(Q, C, \texttt{step} = \text{Reformulate})$
\IF{$d_1$ == \texttt{execute}}
    \STATE $Q_1, \dots, Q_n \leftarrow \text{Reformulate}(Q)$
\ELSE
    \STATE $Q_1 \leftarrow Q$
    \STATE $n \leftarrow 1$
\ENDIF

\STATE \textbf{// Step 2: Rewrite }
\STATE $d_2 \leftarrow \text{Planner}(Q, C, \texttt{step} = \text{Rewrite})$
\IF{$d_2$ == \texttt{execute}}
    \STATE $C \leftarrow \text{Rewrite}(C)$
\ELSE
    \STATE $C \leftarrow C$
\ENDIF

\STATE \textbf{// Step 3: Answer}
\FOR{$i = 1$ to $n$}
    \STATE $A_i \leftarrow \text{Answer}(Q_i, C)$
\ENDFOR
\STATE $A \leftarrow \text{Aggregate}(A_1, \dots, A_n)$

\STATE \textbf{// Step 4: Review }
\STATE $d_3 \leftarrow \text{Planner}(Q, C, A, \texttt{step} = \text{Review})$
\IF{$d_3$ == \texttt{execute}}
    \STATE $A \leftarrow \text{Review}(A, Q, C)$
\ENDIF

\RETURN $A$
\end{algorithmic}
\end{algorithm}

\subsection{Reasoning Pipeline}
Based on the given temporal document and question, the model follows the strategies provided by the LLM planner to perform step-by-step reasoning and obtain the final answer.
To better illustrate the reasoning process of our proposed approach, we present the pipeline in Algorithm~\ref{alg:temporal_qa_global_rewrite}. First, we initialize the answer variable (lines 1–2), then invoke the Planner to decide whether to decompose the question into sub-questions (lines 4–9). Next, the LLM planner determines whether to rewrite the context into a temporally structured form (lines 12–16). Then each sub-question is answered using the context $C$ (lines 18–20), and the answers are aggregated (line 21). Finally, the model may optionally review and revise the answer (lines 23-25) before returning the final output (line 27).

\section{Experiments}
In this section, we conduct a series of experiments to evaluate the effectiveness of the proposed method and offer a detailed analysis.
\begin{table*}[ht]
\begin{center}
\renewcommand{\arraystretch}{1.2}
\setlength{\tabcolsep}{2.5pt}
\resizebox{1\linewidth}{!}{
\begin{tabular}{l|cc|cc|cc|cc|cc}
\toprule
\multicolumn{1}{l|}{\multirow{3}{*}{\textbf{Model}}}
& \multicolumn{4}{c|}{\textbf{TimeQA}} 
& \multicolumn{4}{c|}{\textbf{TempReason}} 
& \multicolumn{2}{c}{\multirow{2}{*}{\textbf{Average$\uparrow$}}} \\
\cline{2-9}
& \multicolumn{2}{c|}{\textbf{Easy-mode}} 
& \multicolumn{2}{c|}{\textbf{Hard-mode}} 
& \multicolumn{2}{c|}{\textbf{OBQA-L2}} 
& \multicolumn{2}{c|}{\textbf{OBQA-L3}} 
& & \\
& EM(\%) & F1(\%)  & EM(\%)  & F1(\%)  & EM(\%)  & F1(\%)  & EM(\%)  & F1(\%)  & EM(\%)  & F1(\%)  \\
\midrule
T5-base$^{\dag}$  &   60.0 & 68.2 & 55.6 & 64.1 & 26.0 & 45.0 & 23.8 & 41.8& 41.3 & 54.8\\
T5-large$^{\dag}$  &  63.1 & 71.6 & 59.5 & 68.1 & 32.7 & 50.9 & 28.8 & 46.8& 46.0 & 59.3\\
REMEMO-base$^{\dag}$   & 61.4 & 70.4 & 58.2 & 67.3 & 33.6 & 51.6 & 28.5 & 44.9& 45.4 & 58.6\\
REMEMO-large$^{\dag}$  & 63.7 & 72.3 & 60.5 & 69.3 & 37.4 & 54.9 & 33.4 & 49.3& 48.8 & 61.5\\ \hline
GPT-4$^{\dag}$ &71.6 &74.2 &54.6 &57.1 &45.4 &52.5 &43.1 &48.5 & 54.3  & 57.5 \\
QAaP$^{\dag}$ &48.2& 58.3 &39.6 &49.3 &- &- &- &- & -  & - \\
TG-LLM$^{\dag}$ &66.4 &69.1 &63.1 &66.4 &42.4 &52.2 &35.6 &46.9 & 51.9  & 58.7 \\
\midrule
LLaMA-3-8B-ICL &1.1 &3.1 &1.7 &3.6 &3.8 &10.0 &1.8 &10.5 & 2.1  & 6.8  \\
LLaMA-3-8B-CoT &29.7 &32.5 &31.6 &33.8 &18.5 &26.1 &16.5 &24.1 & 24.1 (+22.0↑) & 29.1 (+22.3↑) \\
\cem{LLaMA-3-8B-AdapTime$^*$} &\cem{\textbf{41.5}}& \cem{\textbf{47.2}}& \cem{\textbf{33.3}}& \cem{\textbf{38.6}}& \cem{\textbf{18.7}}& \cem{\textbf{26.8}}& \cem{14.5}&\cem{ 22.5} & \cem{\textbf{27.0 (+24.9↑)}} & \cem{\textbf{33.8 (+27.0↑)}} \\
\midrule
Qwen-3-8B-ICL &67.5 &70.3 &56.9 &60.1 &23.9 &33.9 &23.1 &32.7 & 42.9 & 49.3  \\
Qwen-3-8B-CoT &69.4 &71.0 &62.9 &64.7 &22.6 &30.6 &28.8 &33.4 & 45.9 (+3.0↑) & 49.9 (+0.6↑) \\
\cem{Qwen-3-8B-AdapTime$^*$} &\cem{\textbf{72.7}}& \cem{\textbf{74.1}}&\cem{\textbf{66.5}} &\cem{\textbf{68.2} }&\cem{\textbf{29.1}} &\cem{\textbf{37.9} }&\cem{\textbf{28.8}} &\cem{\textbf{33.8}} & \cem{\textbf{49.3 (+6.4↑) }} & \cem{\textbf{53.5 (+4.2↑) }} \\
\midrule
DeepSeek-V3-ICL &80.8 &82.9 &68.8 &71.6 &45.1 &50.8 &43.6 &48.3 & 59.6 & 63.4 \\
DeepSeek-V3-CoT &85.3 &86.7 &75.6 &77.0 &44.8 &49.1 &47.0 &50.4 &63.2 (+3.6↑) &65.8 (+2.4↑) \\
DeepSeek-V3-Step-back &84.4&	86.0&	76.4&	77.9&	45.8&	50.8&	48.8&	52.3&63.9 (+4.3↑)&66.8 (+3.4↑)
 \\
DeepSeek-V3-Self-refinement &77.6	&80.4&	76.4	&78.4	&44.3&	47.2&	41.1	&42.3&60.1 (+0.5↑)&62.3 (-1.1↓)
 \\
\cem{DeepSeek-V3-AdapTime$^*$ }&\cem{\textbf{85.4}} &\cem{86.6} &\cem{ \textbf{77.7}} &\cem{\textbf{79.2} }&\cem{\textbf{48.0 }}&\cem{\textbf{52.1} }&\cem{\textbf{49.8 }}&\cem{\textbf{53.2} }&\cem{\textbf{65.1 (+5.5↑)}} &\cem{\textbf{67.7 (+4.3↑)}} \\
\bottomrule
\end{tabular}}
\caption{Main results using different models and strategies. We report exact match (EM) and token-level F1 scores. In line with previous work, we randomly sampled 1,000 examples under each task. Results with $^*$ are averaged over three random runs ($p <
0.05$ under t-test). The last two columns show average scores across four tasks, and their absolute improvements over the corresponding ICL baselines. Results with $^{\dag}$
are reported in the original papers.}
\label{table-main-results}
\end{center}
\end{table*}
\subsection{Experimental Setup}
\subsubsection{Datasets}
We evaluate our approach on two widely used temporal question answering datasets. TimeQA~\cite{chen2021time} is a human-annotated benchmark consisting of both easy and hard questions. Easy questions can be answered based on explicitly mentioned temporal expressions in the document, while hard questions require additional reasoning. 
TempReason~\cite{tan-etal-2023-towards} defines multiple levels of temporal understanding. We focus on the more challenging ones: aligning time expressions with events (L2) and reasoning about temporal relations between events (L3). To ensure a balanced distribution across different types of questions, we randomly sample 1,000 test instances for each type.
\subsubsection{Evaluation Metrics}
We use Exact Match (EM) and F1 score to evaluate model performance. EM measures the percentage of predictions that exactly match the ground truth, while F1 captures the token-level overlap between predictions and ground truth, computed as the harmonic mean of precision and recall.
\subsubsection{Baselines}
To better analyze our method and enable a broad comparison, we include two lines of recent temporal reasoning approaches as baselines.
One line enhances temporal reasoning through fine-tuning, manual verification, or external tools.
We introduce a set of embedding-based models~\cite{tan-etal-2023-towards,yang2023once}, including T5-base, T5-large, REMEMO-base, and REMEMO-large, as well as a competitive LLM baseline, GPT-4. The results of T5-base and T5-large are based on vanilla checkpoints without any continual pretraining on temporal data, using only standard supervised fine-tuning.~\cite{tan-etal-2023-towards,yang2023once} We also include QAaP~\cite{zhu2023question} and TG-LLM~\cite{xiong2024large}, the previous state-of-the-art methods for temporal reasoning.
The other explores the reasoning capabilities of LLMs using prompting-based techniques in-context learning (ICL) and chain-of-thought (CoT)~\cite{wei2022chain}. Step-back~\cite{zhengtake} is a prompting technique that encourages LLMs to perform higher-level abstraction before answering. Self-Refinement~\cite{madaan2023self} is an agentic method that enables the model to reflect on its initial output and iteratively revise it.
\subsubsection{Implementation Details}
We adopt three competitive open-source LLMs as backbones: LLaMA-3.1-8B-Instruct, Qwen-3-8B, and DeepSeek-V3-0324. The reasoning process only relies on the inherent capabilities of the models, without incorporating any external tools, fine-tuning, or manual corrections. During inference, we adopt a decoding strategy with top-$k$ = 10 and temperature of 0.7. We set the batch size to 1 for single-instance evaluation and limit the output to a maximum of 512 new tokens. 
The prompt templates used are provided in Appendix~\ref{app:prompt}.

\subsection{Main Results}


The experimental results in Table~\ref{table-main-results} show the effectiveness of our method AdapTime.
 We have the following observations:
\begin{itemize}[leftmargin=*, nolistsep]

\item 
AdapTime consistently outperforms both ICL and generalized CoT. ICL often fails to capture complex temporal dependencies, as it lacks explicit reasoning mechanisms, and CoT may generate overly generic reasoning paths that are not well-adapted to temporal questions. In contrast, AdapTime leverages structured decomposition to produce more targeted and accurate inference.
\item Adaptime consistently improves performance across LLMs of different structures or scales, including LLaMA-3-8B, Qwen-3-8B, and DeepSeek-V3. This indicates that the method is model-agnostic and exhibits strong generalizability,~making it applicable to a wide range of language models. Moreover, the gains come from our proposed framework rather than model-specific capabilities.~It achieves better performance on DeepSeek-V3 compared to the previous state-of-the-art method.~Qwen-3-8B equipped with AdapTime even surpass the larger closed-source model GPT-4 on TimeQA-Easy/Hard.
\item AdapTime shows greater improvements on more challenging benchmarks, especially those requiring multi-hop or temporally complex reasoning, since it explicitly guides the model through intermediate reasoning steps.
This highlights its strength in handling difficult question types that involve complex temporal structures.

\end{itemize}

\subsection{Ablation Study}
To assess the effectiveness of our method and each component, we conduct a detailed ablation study on DeepSeek-V3, as shown in Table~\ref{table-ablation-results}. 

First, we compare the baseline ICL (i.e., in-context learning) with three variants that incorporate our proposed reasoning actions: \textit{reformulate},  \textit{rewrite}, and  \textit{review}. Each action leads to consistent performance improvements.
In particular, the rewrite step contributes the most to performance gains, especially in the TimeQA-Easy and OBQA-L2 datasets, highlighting the effectiveness of transforming text into a time-sensitive format.

Second, we removed one action at a time from AdapTime. The removal of any single component results in a slight performance drop, confirming that each action plays an important role in the reasoning process. 
Notably, removing the rewrite module leads to the most significant performance drop over these datasets, suggesting its critical role in aligning temporal context across tasks.

Finally, we eliminated the LLM planner and forced the model to follow a fixed sequence of reasoning steps. 
Performance drops consistently across all settings except TimeQA-Easy without the LLM planner. 
This indicates that adopting appropriate reasoning steps is crucial, as different tasks may require different reasoning processes.







\begin{table}[t]
\small
\setlength\tabcolsep{1.5 pt}  
\begin{center}
\renewcommand{\arraystretch}{1.3}
\resizebox{1\linewidth}{!}{
\begin{tabular}{l|cc|cc|cc|cc}
\toprule
\multicolumn{1}{l|}{\multirow{3}{*}{\textbf{Model}}}
& \multicolumn{4}{c|}{\textbf{TimeQA}} 
& \multicolumn{4}{c}{\textbf{TempReason}} 
\\
\cline{2-9}
& \multicolumn{2}{c|}{\textbf{Easy-mode}} 
& \multicolumn{2}{c|}{\textbf{Hard-mode}} 
& \multicolumn{2}{c|}{\textbf{OBQA-L2}} 
& \multicolumn{2}{c}{\textbf{OBQA-L3}} 
\\
& EM & F1 & EM & F1 & EM & F1 & EM & F1  \\

\midrule
ICL &80.8 &82.9 &68.8 &71.6 &45.1 &50.8 &43.6 &48.3  \\
w/ reform. &84.3 &85.8 &77.0 &78.9 &46.7 &50.8 &49.3 &51.9  \\
w/ rewrite &\textbf{86.4} &\textbf{87.6} &76.0 &77.8 &46.9 &52.1 &48.5 &51.9  \\
w/ review &85.8 &86.9 &75.9 &77.4 &46.8 &50.6 &48.1 &51.4  \\ 
\midrule
w/o reform. &85	&86.6	&76.8&	78.5
 &46.9 &51.8 &48.9	&51.8  \\
w/o rewrite &84.8	&86.5&	76.5&	77.9
 &46.1 &49.8 & 47.6	&50.7  \\
w/o review &84.8&	86.2&	77.2&	78.7
 &47.0 &51.2 & 49.0	&51.6  \\
 w/o Planner&85.3	&86.7&	77.1&	78.8&	47.1&	51.3&	48.9&	52.0
\\
AdapTime &85.4 &86.6 &\textbf{77.7} &\textbf{79.2} &\textbf{48.0 }&\textbf{52.1} &\textbf{49.8 }&\textbf{53.2}  \\
\bottomrule
\end{tabular}}
\caption{Ablation study on DeepSeek-V3. All metrics are reported as percentages.}
\label{table-ablation-results}
\end{center}
\end{table}

\subsection{Computational Cost Analysis}
To quantify cost, we report the average total number of tokens (input + output) per instance across methods on the TempReason-L3 dataset in Table~\ref{tab:cost}. Although AdapTime uses slightly more tokens due to the inclusion of reasoning instructions and intermediate steps, the increase is modest and much smaller than that of iterative or multi-call methods such as self-refinement. Moreover, since the temporal context is often lengthy, the majority of token cost comes from the input text, which constitutes a fixed overhead across methods. 
\begin{table}[t]
\centering
\setlength\tabcolsep{9.pt}
\resizebox{0.4\textwidth}{!}{
\begin{tabular}{lc}
\toprule
               &Avg. Total Tokens    \\ \hline
ICL &    4345.04	
      \\
CoT &  4584.17	
 \\
Self-refinement& $>10000	$
 	
 \\ \hline
AdapTime   & 4873.64
  \\ \bottomrule
\end{tabular}}
\caption{Computational Cost Analysis.}
\label{tab:cost}
\vspace{-0.3cm}
\end{table}

\subsection{Planner Comparison}

To further investigate planner quality, we conducted an experiment where we sampled 1,000 high-quality planning trajectories from DeepSeek-V3 outputs on the training set, and used them to fine-tune a LLaMA-3-8B model as a supervised planner. Surprisingly, the fine-tuned planner underperformed compared to our original prompt-based adaptive planner as shown in Table \ref{tab:planner1}. These results suggest that fine-tuning a planner may introduce overfitting or poor generalization. In contrast, our prompt-based planner leverages the in-context reasoning capability of strong LLMs, offering better generalization and flexibility.
\begin{table}[t]
\centering
\setlength\tabcolsep{5.pt}
\renewcommand{\arraystretch}{1}
\resizebox{1\linewidth}{!}{
\begin{tabular}{l|cc|cc}
\toprule
\multirow{2}{*}{\textbf{Method}} 

& \multicolumn{2}{c|}{\textbf{TimeQA-Easy}} & \multicolumn{2}{c}{\textbf{TimeQA-Hard}} \\
& EM(\%) & F1(\%) & EM (\%)& F1(\%) \\
\midrule
CoT &29.7& 32.5& 31.6 &33.8  \\
AdapTime-Finetuned &31.0& 40.5& 23.4& 32.0 

 \\
AdapTime(Ours) &\textbf{41.5} &\textbf{47.2}& \textbf{33.3}& \textbf{38.6} 
 \\
\bottomrule
\end{tabular}}
\caption{Performance of different planner. We use LLaMA-3-8B as the backbone model.}
\label{tab:planner1}
\end{table}

\subsection{Rule Distribution}
\begin{figure}[t]
\centering
\resizebox{1\linewidth}{!}{
\begin{tikzpicture}
\begin{axis}[
    enlargelimits=0.15,
    symbolic x coords={Reformulate, Rewrite, Review},
    xtick=data,
    enlargelimits=0.3,
    xticklabel style={font=\small},
    tick label style={/pgf/number format/1000 sep=\,},
    ybar,
    yticklabel=\pgfmathprintnumber{\tick}\%,
    bar width=16pt,
    width=9.1cm,
    height=5cm,
    nodes near coords,
    nodes near coords align={vertical},
    nodes near coords style={font=\scriptsize, yshift=0pt},
    font=\small,
    grid=major,
    clip=true,
    legend style={
        draw=none,
        fill=none,
        at={(0.5,-0.25)},
        anchor=north,
        legend columns=2
    },
]

\addplot[
    fill=orange!60!white,
    draw=orange,
    bar shift=-24pt
] coordinates {
    (Reformulate, 80.1)
    (Rewrite, 86.4)
    (Review, 33.4)
};

\addplot[
    fill=gray!60!black,
    draw=black,
    bar shift=-8pt
] coordinates {
    (Reformulate, 69.3)
    (Rewrite, 88.1)
    (Review, 40.4)
};

\addplot[
    fill=black!30!white,
    draw=black,
    bar shift=8pt
] coordinates {
    (Reformulate, 34.2)
    (Rewrite, 85.8)
    (Review, 56.2)
};

\addplot[
    fill=brown!40!white,
    draw=brown!60!black,
    bar shift=24pt
] coordinates {
    (Reformulate, 26.7)
    (Rewrite, 88.2)
    (Review, 72.5)
};

\legend{TimeQA-Easy, TimeQA-Hard, TempReason-L2, TempReason-L3}
\end{axis}
\end{tikzpicture}}
\caption{Comparison of the proportion of each operation in different datasets.}
\label{action}
\end{figure}
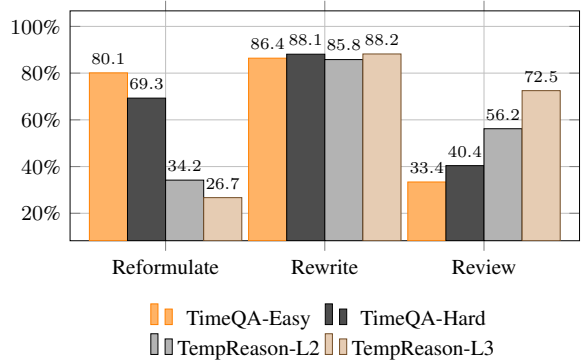

We hypothesize that different types of temporal questions require distinct reasoning steps. In our ablation studies, we have already demonstrated the effectiveness of using an LLM planner to define temporal reasoning strategy. To further assess the motivation of the action combination generated by the planner and to understand what types of steps it produces for different question types, we analyzed the action distributions across multiple datasets.
The results in Figure~\ref{action} show that our method is capable of adaptively choosing reasoning actions:
\begin{itemize}[leftmargin=*,nolistsep]
\item For questions with clear structure, the model performs decomposition to ensure that each sub-question is simple and unambiguous. For example, in TimeQA, questions often involve events at a specific time point. These can be decomposed into: \{(1) What are the possible answers regardless of time? and (2) Which answer holds true at the given time?\}
In contrast, for less clearly structured questions, the model adopts a more conservative strategy: it first reconstructs the timeline to derive the answer, and then verifies the result. 
For instance, in TempReason, L2 and L3 questions require flexible multi-step inference over temporal spans and event order.
\item In all settings, the rewrite action is frequently chosen. This confirms that rewriting natural language context into a time-sensitive format is essential for effective temporal reasoning. 

\item On more challenging datasets such as TempReason-L2/L3, the model tends to be less confident in its initial predictions and therefore prefers to select the review action as a form of self-verification.
\end{itemize}
Overall, the results of our action distribution analysis confirm that the LLM planner generates appropriate reasoning strategies tailored to different types of temporal questions, which in turn improves overall model performance.
\subsection{Open-domain Temporal Reasoning}
We further evaluate AdapTime on an open-domain setting using 1,000 questions from the ArchivalQA benchmark~\cite{wang2022archivalqa}. 
As shown in Table~\ref{tab:open-domain}, AdapTime significantly outperforms the baseline methods in both accuracy and F1.

Our method is modular and can be combined with retrievers such as BM25. While retrieval quality would affect planner decisions and downstream reasoning~\cite{align2025,harness2025}, AdapTime’s adaptive mechanism remains effective in handling noisy or incomplete evidence through selective verification and timeline reconstruction. We applied BM25 on TempReason-L3, retrieving the most relevant sentences as context. We then applied AdapTime using these retrieved contexts. Results in \ref{tab:rag} show that AdapTime consistently improves reasoning performance over retrieved content. This confirms that AdapTime is compatible with retrievers and can improve performance in retrieval-augmented, open-domain settings. It also shows that our method remains effective when operating on retrieved, potentially noisy input, rather than clean gold contexts.
\begin{table}[t]
\centering
    \setlength\tabcolsep{15.pt}
\resizebox{0.4\textwidth}{!}{
\begin{tabular}{lcc}
\toprule
               &ACC&F1    \\ \hline
Deepseek-V3 &  19.7  &18.6	
      \\
Step-back & 29.5 &27.8	
 \\
Self-refinement&28.2&24.3
 	
 \\ \hline
AdapTime   & \textbf{32.2}&\textbf{30.5}
  \\ \bottomrule
\end{tabular}}
\caption{Experimental results on the open-domain temporal QA benchmark ArchivalQA.}
\label{tab:open-domain}
\vspace{-0.3cm}
\end{table}
\begin{table}[t]
\centering
    \setlength\tabcolsep{15.pt}
\resizebox{0.4\textwidth}{!}{
\begin{tabular}{lcc}
\toprule
               &EM&F1    \\ \hline
BM25 &   44.7&	47.7

 \\ 
BM25+AdapTime   & \textbf{46.7}&	\textbf{49.9}
  \\ \bottomrule
\end{tabular}}
\caption{Performance with retriever integration on the TempReason-l3 dataset.}
\label{tab:rag}
\vspace{-0.3cm}
\end{table}


\subsection{Case Study}
We conducted a case study to analyze the reasoning trajectories generated by our method. 
Unlike standard prompting methods that often struggle with implicit temporal structures, AdapTime explicitly guides the model through intermediate reasoning stages, enhancing its ability to handle chronological dependencies and integrate commonsense knowledge with textual evidence. This highlights the model’s capacity for adaptive and interpretable reasoning, which is particularly valuable in temporally rich QA scenarios. A detailed example is provided in the Appendix~\ref{app:case}, illustrating how our model performs reasoning and demonstrating its superior ability to interpret temporal information compared to other methods.

\section{Conclusion}
In this paper, we propose a novel temporal reasoning approach AdapTime that enables LLMs to adaptively perform time-aware inference.
AdapTime consistently outperforms strong baselines, exhibiting robust generalizability and significant gains on temporally complex tasks.~Moreover, it can be seamlessly integrated with state-of-the-art LLMs. In the future, we plan to incorporate more temporal actions and planners to further enhance the model's capability for robust temporal reasoning.

\section*{Limitations}
AdapTime addresses the limitations of existing methods, which often rely on external support or rigid reasoning pipelines and struggle to generalize across diverse temporal tasks. However, relying solely on LLMs as planners can be unstable in practice, leading to inconsistent planning outcomes across runs or tasks. as their behavior may exhibit a degree of randomness. Moreover, the current action set can be further expanded to include more fine-grained temporal operations or practical tools to improve precision and applicability. In future work, we plan to explore more temporal actions, integrate symbolic or external components, and develop techniques to improve the reliability and controllability of LLM-based planners.

\section*{Acknowledgments}
This work is in part funded by the National Key Research and Development
Program of China (2023YFC3321600); in part by National Natural Science Foundation of China (Grant No. 62372364) and the Technical Innovation Guidance Plan of Shaanxi Province, China (Grant No. 2024QCY-KXJ-199); in part by  National Natural Science Foundation of China (No.62502404), Hong Kong Research Grants Council (Research Impact Fund No.R1015-23, Collaborative Research Fund No.C1043-24GF, General Research Fund No. 11218325), Institute of Digital Medicine of City University of Hong Kong (No.9229503), Huawei (Huawei Innovation Research Program), Tencent (Tencent Rhino-Bird Focused Research Program, Tencent University Cooperation Project), Didi (CCF-Didi Gaia Scholars Research Fund), Kuaishou (CCF-Kuaishou Large Model Explorer Fund No. 2025008, Kuaishou University Cooperation Project), and Bytedance.

\bibliography{main}

@inproceedings{son2023time,
  title={Time-Aware Representation Learning for Time-Sensitive Question Answering},
  author={Son, Jungbin and Oh, Alice},
  booktitle={Findings of the Association for Computational Linguistics: EMNLP 2023},
  pages={70--77},
  year={2023}
}

@inproceedings{zhu2023question,
  title={Question Answering as Programming for Solving Time-Sensitive Questions},
  author={Zhu, Xinyu and Yang, Cheng and Chen, Bei and Li, Siheng and Lou, Jian-Guang and Yang, Yujiu},
  booktitle={Proceedings of the 2023 Conference on Empirical Methods in Natural Language Processing},
  pages={12775--12790},
  year={2023}
}

@inproceedings{yin2024time,
  title={Time-CoT for Enhancing Time Reasoning Factual Question Answering in Large Language Models},
  author={Yin, Baosheng and Hu, Naiyu},
  booktitle={2024 International Joint Conference on Neural Networks (IJCNN)},
  pages={1--8},
  year={2024},
  organization={IEEE}
}

@article{xiong2024large,
  title={Large language models can learn temporal reasoning},
  author={Xiong, Siheng and Payani, Ali and Kompella, Ramana and Fekri, Faramarz},
  journal={arXiv preprint arXiv:2401.06853},
  year={2024}
}

@inproceedings{wu2024event,
  title={An Event-based Abductive Learning for Hard Time-sensitive Question Answering},
  author={Wu, Shaojuan and Li, Jitong and Zhang, Xiaowang and Feng, Zhiyong},
  booktitle={Proceedings of the 2024 Joint International Conference on Computational Linguistics, Language Resources and Evaluation (LREC-COLING 2024)},
  pages={1105--1115},
  year={2024}
}

@inproceedings{chen2021time,
  title     = {A Dataset for Answering Time-Sensitive Questions},
  author    = {Chen, Wenhu and Wang, Xinyi and Wang, William Yang},
  booktitle = {Proceedings of the Neural Information Processing Systems Track on Datasets and Benchmarks (NeurIPS Datasets and Benchmarks)},
  year      = {2021}
}

@inproceedings{tan-etal-2023-towards,
    title = "Towards Benchmarking and Improving the Temporal Reasoning Capability of Large Language Models",
    author = "Tan, Qingyu  and
      Ng, Hwee Tou  and
      Bing, Lidong",
    editor = "Rogers, Anna  and
      Boyd-Graber, Jordan  and
      Okazaki, Naoaki",
    booktitle = "Proceedings of the 61st Annual Meeting of the Association for Computational Linguistics (Volume 1: Long Papers)",
    month = jul,
    year = "2023",
    address = "Toronto, Canada",
    publisher = "Association for Computational Linguistics",
    url = "https://aclanthology.org/2023.acl-long.828/",
    doi = "10.18653/v1/2023.acl-long.828",
    pages = "14820--14835"
}

@inproceedings{shao2023prompting,
  title={Prompting large language models with answer heuristics for knowledge-based visual question answering},
  author={Shao, Zhenwei and Yu, Zhou and Wang, Meng and Yu, Jun},
  booktitle={Proceedings of the IEEE/CVF Conference on computer vision and pattern recognition},
  pages={14974--14983},
  year={2023}
}

@inproceedings{kamalloo2023evaluating,
  title={Evaluating Open-Domain Question Answering in the Era of Large Language Models},
  author={Kamalloo, Ehsan and Dziri, Nouha and Clarke, Charles and Rafiei, Davood},
  booktitle={Proceedings of the 61st Annual Meeting of the Association for Computational Linguistics (Volume 1: Long Papers)},
  pages={5591--5606},
  year={2023}
}

@article{dong2024cost,
  title={Cost-efficient knowledge-based question answering with large language models},
  author={Dong, Junnan and Zhang, Qinggang and Zhou, Chuang and Chen, Hao and Zha, Daochen and Huang, Xiao},
  journal={Advances in Neural Information Processing Systems},
  volume={37},
  pages={115261--115281},
  year={2024}
}

@inproceedings{zhengtake,
  title={Take a Step Back: Evoking Reasoning via Abstraction in Large Language Models},
  author={Zheng, Huaixiu Steven and Mishra, Swaroop and Chen, Xinyun and Cheng, Heng-Tze and Chi, Ed H and Le, Quoc V and Zhou, Denny},
  booktitle={The Twelfth International Conference on Learning Representations},
  year={2024}
}

@article{wei2022chain,
  title={Chain-of-thought prompting elicits reasoning in large language models},
  author={Wei, Jason and Wang, Xuezhi and Schuurmans, Dale and Bosma, Maarten and Xia, Fei and Chi, Ed and Le, Quoc V and Zhou, Denny and others},
  journal={Advances in neural information processing systems},
  volume={35},
  pages={24824--24837},
  year={2022}
}

@inproceedings{wang2023can,
  title={Can ChatGPT Defend its Belief in Truth? Evaluating LLM Reasoning via Debate},
  author={Wang, Boshi and Yue, Xiang and Sun, Huan},
  booktitle={Findings of the Association for Computational Linguistics: EMNLP 2023},
  pages={11865--11881},
  year={2023}
}

@inproceedings{laban2023summedits,
  title={SummEdits: Measuring LLM ability at factual reasoning through the lens of summarization},
  author={Laban, Philippe and Kry{\'s}ci{\'n}ski, Wojciech and Agarwal, Divyansh and Fabbri, Alexander Richard and Xiong, Caiming and Joty, Shafiq and Wu, Chien-Sheng},
  booktitle={Proceedings of the 2023 conference on empirical methods in natural language processing},
  pages={9662--9676},
  year={2023}
}

@inproceedings{qiu2024snapntell,
  title={SnapNTell: Enhancing Entity-Centric Visual Question Answering with Retrieval Augmented Multimodal LLM},
  author={Qiu, Jielin and Madotto, Andrea and Lin, Zhaojiang and Crook, Paul A and Xu, Yifan and Damavandi, Babak and Dong, Xin Luna and Faloutsos, Christos and Li, Lei and Moon, Seungwhan},
  booktitle={Findings of the Association for Computational Linguistics: EMNLP 2024},
  pages={247--266},
  year={2024}
}

@inproceedings{li2023few,
  title={Few-shot In-context Learning on Knowledge Base Question Answering},
  author={Li, Tianle and Ma, Xueguang and Zhuang, Alex and Gu, Yu and Su, Yu and Chen, Wenhu},
  booktitle={Proceedings of the 61st Annual Meeting of the Association for Computational Linguistics (Volume 1: Long Papers)},
  pages={6966--6980},
  year={2023}
}

@inproceedings{dong2024survey,
  title={A Survey on In-context Learning},
  author={Dong, Qingxiu and Li, Lei and Dai, Damai and Zheng, Ce and Ma, Jingyuan and Li, Rui and Xia, Heming and Xu, Jingjing and Wu, Zhiyong and Chang, Baobao and others},
  booktitle={Proceedings of the 2024 Conference on Empirical Methods in Natural Language Processing},
  pages={1107--1128},
  year={2024}
}

@inproceedings{li2024flexkbqa,
  title={Flexkbqa: A flexible llm-powered framework for few-shot knowledge base question answering},
  author={Li, Zhenyu and Fan, Sunqi and Gu, Yu and Li, Xiuxing and Duan, Zhichao and Dong, Bowen and Liu, Ning and Wang, Jianyong},
  booktitle={Proceedings of the AAAI conference on artificial intelligence},
  volume={38},
  number={17},
  pages={18608--18616},
  year={2024}
}

@article{stechly2024chain,
  title={Chain of thoughtlessness? an analysis of cot in planning},
  author={Stechly, Kaya and Valmeekam, Karthik and Kambhampati, Subbarao},
  journal={Advances in Neural Information Processing Systems},
  volume={37},
  pages={29106--29141},
  year={2024}
}

@inproceedings{yeo2025demystifying,
  title={Demystifying Long Chain-of-Thought Reasoning in LLMs},
  author={Yeo, Edward and Tong, Yuxuan and Niu, Xinyao and Neubig, Graham and Yue, Xiang},
  booktitle={ICLR 2025 Workshop on Deep Generative Model in Machine Learning: Theory, Principle and Efficacy}
}

@article{chu2023timebench,
  title={Timebench: A comprehensive evaluation of temporal reasoning abilities in large language models},
  author={Chu, Zheng and Chen, Jingchang and Chen, Qianglong and Yu, Weijiang and Wang, Haotian and Liu, Ming and Qin, Bing},
  journal={arXiv preprint arXiv:2311.17667},
  year={2023}
}

@inproceedings{yang2023once,
  title={Once Upon a Time in Graph: Relative-Time Pretraining for Complex Temporal Reasoning},
  author={Yang, Sen and Li, Xin and Bing, Lidong and Lam, Wai},
  booktitle={Proceedings of the 2023 Conference on Empirical Methods in Natural Language Processing},
  pages={11879--11895},
  year={2023}
}

@article{bazaga2025learning,
  title={Learning to Reason Over Time: Timeline Self-Reflection for Improved Temporal Reasoning in Language Models},
  author={Bazaga, Adri{\'a}n and Blloshmi, Rexhina and Byrne, Bill and de Gispert, Adri{\`a}},
  journal={arXiv preprint arXiv:2504.05258},
  year={2025}
}

@inproceedings{wang2022archivalqa,
  title={Archivalqa: A large-scale benchmark dataset for open-domain question answering over historical news collections},
  author={Wang, Jiexin and Jatowt, Adam and Yoshikawa, Masatoshi},
  booktitle={Proceedings of the 45th International ACM SIGIR Conference on Research and Development in Information Retrieval},
  pages={3025--3035},
  year={2022}
}

@article{madaan2023self,
  title={Self-refine: Iterative refinement with self-feedback},
  author={Madaan, Aman and Tandon, Niket and Gupta, Prakhar and Hallinan, Skyler and Gao, Luyu and Wiegreffe, Sarah and Alon, Uri and Dziri, Nouha and Prabhumoye, Shrimai and Yang, Yiming and others},
  journal={Advances in Neural Information Processing Systems},
  volume={36},
  pages={46534--46594},
  year={2023}
}

@inproceedings{
zhang2026evoking,
title={Evoking User Memory: Personalizing {LLM} via Recollection-Familiarity Adaptive Retrieval},
author={Yingyi Zhang and Junyi Li and Wenlin Zhang and Pengyue Jia and Xianneng Li and Yichao Wang and Derong Xu and Yi Wen and Huifeng Guo and Yong Liu and Xiangyu Zhao},
booktitle={The Fourteenth International Conference on Learning Representations},
year={2026},
url={https://openreview.net/forum?id=f7p0F2X6XN}
}

@inproceedings{bridging2025,
  author       = {Pengyue Jia and
                  Derong Xu and
                  Xiaopeng Li and
                  Zhaocheng Du and
                  Xiangyang Li and
                  Yichao Wang and
                  Yuhao Wang and
                  Qidong Liu and
                  Maolin Wang and
                  Huifeng Guo and
                  Ruiming Tang and
                  Xiangyu Zhao},
  editor       = {Wanxiang Che and
                  Joyce Nabende and
                  Ekaterina Shutova and
                  Mohammad Taher Pilehvar},
  title        = {Bridging Relevance and Reasoning: Rationale Distillation in Retrieval-Augmented
                  Generation},
  booktitle    = {Findings of the Association for Computational Linguistics, {ACL} 2025,
                  Vienna, Austria, July 27 - August 1, 2025},
  series       = {Findings of {ACL}},
  pages        = {4242--4256},
  publisher    = {Association for Computational Linguistics},
  year         = {2025},
  url          = {https://aclanthology.org/2025.findings-acl.220/},
  bibsource    = {dblp computer science bibliography, https://dblp.org}
}

@article{navigate2025,
  author       = {Jingtong Gao and
                  Ling Pan and
                  Yejing Wang and
                  Rui Zhong and
                  Chi Lu and
                  Qingpeng Cai and
                  Peng Jiang and
                  Xiangyu Zhao},
  title        = {Navigate the Unknown: Enhancing {LLM} Reasoning with Intrinsic Motivation
                  Guided Exploration},
  journal      = {CoRR},
  volume       = {abs/2505.17621},
  year         = {2025},
  url          = {https://doi.org/10.48550/arXiv.2505.17621},
  doi          = {10.48550/ARXIV.2505.17621},
  eprinttype   = {arXiv},
  eprint       = {2505.17621},
  bibsource    = {dblp computer science bibliography, https://dblp.org}
}

@inproceedings{stepwise2025,
  author       = {Jingyu Peng and
                  Maolin Wang and
                  Xiangyu Zhao and
                  Kai Zhang and
                  Wanyu Wang and
                  Pengyue Jia and
                  Qidong Liu and
                  Ruocheng Guo and
                  Qi Liu},
  editor       = {Wanxiang Che and
                  Joyce Nabende and
                  Ekaterina Shutova and
                  Mohammad Taher Pilehvar},
  title        = {Stepwise Reasoning Disruption Attack of LLMs},
  booktitle    = {Proceedings of the 63rd Annual Meeting of the Association for Computational
                  Linguistics (Volume 1: Long Papers), {ACL} 2025, Vienna, Austria,
                  July 27 - August 1, 2025},
  pages        = {5040--5058},
  publisher    = {Association for Computational Linguistics},
  year         = {2025},
  url          = {https://aclanthology.org/2025.acl-long.251/},
  bibsource    = {dblp computer science bibliography, https://dblp.org}
}

@article{align2025,
  author       = {Derong Xu and
                  Pengyue Jia and
                  Xiaopeng Li and
                  Yingyi Zhang and
                  Maolin Wang and
                  Qidong Liu and
                  Xiangyu Zhao and
                  Yichao Wang and
                  Huifeng Guo and
                  Ruiming Tang and
                  Enhong Chen and
                  Tong Xu},
  title        = {Align-GRAG: Reasoning-Guided Dual Alignment for Graph Retrieval-Augmented
                  Generation},
  journal      = {CoRR},
  volume       = {abs/2505.16237},
  year         = {2025},
  url          = {https://doi.org/10.48550/arXiv.2505.16237},
  doi          = {10.48550/ARXIV.2505.16237},
  eprinttype   = {arXiv},
  eprint       = {2505.16237},
  bibsource    = {dblp computer science bibliography, https://dblp.org}
}

@inproceedings{llm4rank2025,
  author       = {Jingtong Gao and
                  Bo Chen and
                  Xiangyu Zhao and
                  Weiwen Liu and
                  Xiangyang Li and
                  Yichao Wang and
                  Wanyu Wang and
                  Huifeng Guo and
                  Ruiming Tang},
  editor       = {Guodong Long and
                  Michale Blumestein and
                  Yi Chang and
                  Liane Lewin{-}Eytan and
                  Zi Helen Huang and
                  Elad Yom{-}Tov},
  title        = {LLM4Rerank: LLM-based Auto-Reranking Framework for Recommendations},
  booktitle    = {Proceedings of the {ACM} on Web Conference 2025, {WWW} 2025, Sydney,
                  NSW, Australia, 28 April 2025- 2 May 2025},
  pages        = {228--239},
  publisher    = {{ACM}},
  year         = {2025},
  url          = {https://doi.org/10.1145/3696410.3714922},
  doi          = {10.1145/3696410.3714922},
  bibsource    = {dblp computer science bibliography, https://dblp.org}
}

@article{unified2025,
  author       = {Zichuan Fu and
                  Xiangyang Li and
                  Chuhan Wu and
                  Yichao Wang and
                  Kuicai Dong and
                  Xiangyu Zhao and
                  Mengchen Zhao and
                  Huifeng Guo and
                  Ruiming Tang},
  title        = {A Unified Framework for Multi-Domain {CTR} Prediction via Large Language
                  Models},
  journal      = {{ACM} Trans. Inf. Syst.},
  volume       = {43},
  number       = {5},
  pages        = {117:1--117:33},
  year         = {2025},
  url          = {https://doi.org/10.1145/3698878},
  doi          = {10.1145/3698878},
  bibsource    = {dblp computer science bibliography, https://dblp.org}
}

@inproceedings{plate2023,
  author       = {Yuhao Wang and
                  Xiangyu Zhao and
                  Bo Chen and
                  Qidong Liu and
                  Huifeng Guo and
                  Huanshuo Liu and
                  Yichao Wang and
                  Rui Zhang and
                  Ruiming Tang},
  editor       = {Hsin{-}Hsi Chen and
                  Wei{-}Jou (Edward) Duh and
                  Hen{-}Hsen Huang and
                  Makoto P. Kato and
                  Josiane Mothe and
                  Barbara Poblete},
  title        = {{PLATE:} {A} Prompt-Enhanced Paradigm for Multi-Scenario Recommendations},
  booktitle    = {Proceedings of the 46th International {ACM} {SIGIR} Conference on
                  Research and Development in Information Retrieval, {SIGIR} 2023, Taipei,
                  Taiwan, July 23-27, 2023},
  pages        = {1498--1507},
  publisher    = {{ACM}},
  year         = {2023},
  url          = {https://doi.org/10.1145/3539618.3591750},
  doi          = {10.1145/3539618.3591750},
  bibsource    = {dblp computer science bibliography, https://dblp.org}
}

@inproceedings{llmemb2025,
  author       = {Qidong Liu and
                  Xian Wu and
                  Wanyu Wang and
                  Yejing Wang and
                  Yuanshao Zhu and
                  Xiangyu Zhao and
                  Feng Tian and
                  Yefeng Zheng},
  editor       = {Toby Walsh and
                  Julie Shah and
                  Zico Kolter},
  title        = {LLMEmb: Large Language Model Can Be a Good Embedding Generator for
                  Sequential Recommendation},
  booktitle    = {Thirty-Ninth {AAAI} Conference on Artificial Intelligence, Thirty-Seventh
                  Conference on Innovative Applications of Artificial Intelligence,
                  Fifteenth Symposium on Educational Advances in Artificial Intelligence,
                  {AAAI} 2025, Philadelphia, PA, USA, February 25 - March 4, 2025},
  pages        = {12183--12191},
  publisher    = {{AAAI} Press},
  year         = {2025},
  url          = {https://doi.org/10.1609/aaai.v39i11.33327},
  doi          = {10.1609/AAAI.V39I11.33327},
  bibsource    = {dblp computer science bibliography, https://dblp.org}
}

@inproceedings{large2025,
  author       = {Qidong Liu and
                  Xiangyu Zhao and
                  Yuhao Wang and
                  Yejing Wang and
                  Zijian Zhang and
                  Yuqi Sun and
                  Xiang Li and
                  Maolin Wang and
                  Pengyue Jia and
                  Chong Chen and
                  Wei Huang and
                  Feng Tian},
  editor       = {Luiza Antonie and
                  Jian Pei and
                  Xiaohui Yu and
                  Flavio Chierichetti and
                  Hady W. Lauw and
                  Yizhou Sun and
                  Srinivasan Parthasarathy},
  title        = {Large Language Model Enhanced Recommender Systems: Methods, Applications
                  and Trends},
  booktitle    = {Proceedings of the 31st {ACM} {SIGKDD} Conference on Knowledge Discovery
                  and Data Mining, V.2, {KDD} 2025, Toronto ON, Canada, August 3-7,
                  2025},
  pages        = {6096--6106},
  publisher    = {{ACM}},
  year         = {2025},
  url          = {https://doi.org/10.1145/3711896.3736553},
  doi          = {10.1145/3711896.3736553},
  bibsource    = {dblp computer science bibliography, https://dblp.org}
}

@inproceedings{rethinking2025,
  author       = {Hanbing Wang and
                  Xiaorui Liu and
                  Wenqi Fan and
                  Xiangyu Zhao and
                  Venkataramana Kini and
                  Devendra Pratap Yadav and
                  Fei Wang and
                  Zhen Wen and
                  Hui Liu},
  editor       = {Kentaro Inui and
                  Sakriani Sakti and
                  Haofen Wang and
                  Derek F. Wong and
                  Pushpak Bhattacharyya and
                  Biplab Banerjee and
                  Asif Ekbal and
                  Tanmoy Chakraborty and
                  Dhirendra Pratap Singh},
  title        = {Rethinking Large Language Model Architectures for Sequential Recommendations},
  booktitle    = {Proceedings of the 14th International Joint Conference on Natural
                  Language Processing and the 4th Conference of the Asia-Pacific Chapter
                  of the Association for Computational Linguistics, {IJCNLP-AACL} 2025,
                  Mumbai, India, December 20-24, 2025},
  pages        = {3376--3391},
  publisher    = {The Asian Federation of Natural Language Processing and The Association
                  for Computational Linguistics},
  year         = {2025},
  url          = {https://aclanthology.org/2025.ijcnlp-long.180/},
  bibsource    = {dblp computer science bibliography, https://dblp.org}
}

@inproceedings{harness2025,
  author       = {Derong Xu and
                  Xinhang Li and
                  Ziheng Zhang and
                  Zhenxi Lin and
                  Zhihong Zhu and
                  Zhi Zheng and
                  Xian Wu and
                  Xiangyu Zhao and
                  Tong Xu and
                  Enhong Chen},
  editor       = {Toby Walsh and
                  Julie Shah and
                  Zico Kolter},
  title        = {Harnessing Large Language Models for Knowledge Graph Question Answering
                  via Adaptive Multi-Aspect Retrieval-Augmentation},
  booktitle    = {Thirty-Ninth {AAAI} Conference on Artificial Intelligence, Thirty-Seventh
                  Conference on Innovative Applications of Artificial Intelligence,
                  Fifteenth Symposium on Educational Advances in Artificial Intelligence,
                  {AAAI} 2025, Philadelphia, PA, USA, February 25 - March 4, 2025},
  pages        = {25570--25578},
  publisher    = {{AAAI} Press},
  year         = {2025},
  url          = {https://doi.org/10.1609/aaai.v39i24.34747},
  doi          = {10.1609/AAAI.V39I24.34747},
  bibsource    = {dblp computer science bibliography, https://dblp.org}
}

@inproceedings{mill2024a,
  author       = {Pengyue Jia and
                  Yiding Liu and
                  Xiangyu Zhao and
                  Xiaopeng Li and
                  Changying Hao and
                  Shuaiqiang Wang and
                  Dawei Yin},
  editor       = {Kevin Duh and
                  Helena G{\'{o}}mez{-}Adorno and
                  Steven Bethard},
  title        = {{MILL:} Mutual Verification with Large Language Models for Zero-Shot
                  Query Expansion},
  booktitle    = {Proceedings of the 2024 Conference of the North American Chapter of
                  the Association for Computational Linguistics: Human Language Technologies
                  (Volume 1: Long Papers), {NAACL} 2024, Mexico City, Mexico, June 16-21,
                  2024},
  pages        = {2498--2518},
  publisher    = {Association for Computational Linguistics},
  year         = {2024},
  url          = {https://doi.org/10.18653/v1/2024.naacl-long.138},
  doi          = {10.18653/V1/2024.NAACL-LONG.138},
  bibsource    = {dblp computer science bibliography, https://dblp.org}
}

@inproceedings{mesh,
  author       = {Yimin Deng and
                  Yuxia Wu and
                  Yejing Wang and
                  Guoshuai Zhao and
                  Li Zhu and
                  Qidong Liu and
                  Derong Xu and
                  Zichuan Fu and
                  Xian Wu and
                  Yefeng Zheng and
                  Xiangyu Zhao and
                  Xueming Qian},
  editor       = {Wanxiang Che and
                  Joyce Nabende and
                  Ekaterina Shutova and
                  Mohammad Taher Pilehvar},
  title        = {A Multi-Expert Structural-Semantic Hybrid Framework for Unveiling
                  Historical Patterns in Temporal Knowledge Graphs},
  booktitle    = {Findings of the Association for Computational Linguistics, {ACL} 2025,
                  Vienna, Austria, July 27 - August 1, 2025},
  series       = {Findings of {ACL}},
  pages        = {20553--20565},
  publisher    = {Association for Computational Linguistics},
  year         = {2025},
  url          = {https://aclanthology.org/2025.findings-acl.1056/},
  bibsource    = {dblp computer science bibliography, https://dblp.org}
}
\clearpage
\appendix

\section{Prompt}\label{app:prompt}
In this section, we present the prompt templates used for each type of action in our AdapTime framework. As described in the methods section, AdapTime decomposes the reasoning process into three structured operations: Reformulate, Rewrite, and Review. To guide LLMs to perform these operations in a controlled and interpretable manner, we design simple yet effective prompt templates for each action type. These templates are instantiated dynamically during multi-step reasoning, depending on the needs of each case.

The Reformulate action aims to simplify the original complex question into a set of easier, focused sub-questions that are easier for the model to answer individually. This step encourages the model to disentangle temporal or logical dependencies embedded in the original query, facilitating more accurate downstream reasoning.
\begin{tcolorbox}[colframe=brown,
        width=1\linewidth,
        arc=1mm, 
        auto outer arc,
        title={\small Reformulate Template},
        breakable,]
       Break the question \underline{QUESTION} down into several simple sub-questions and answer each sub-question. Then return the final answer.
\end{tcolorbox}
The Rewrite action prompts the model to explicitly construct a timeline from the narrative context relevant to the question. This is particularly useful for questions that require temporal grounding, such as identifying when an event happened, or what happened at a particular point in time. By asking the model to generate a timeline and align it with the question, we encourage temporal abstraction and normalization of narrative information.
\begin{tcolorbox}[colframe=brown,
        width=1\linewidth,
        arc=1mm, 
        auto outer arc,
        title={\small Rewrite Template},
        breakable,]
       In the context of \underline{STORY} and the \underline{QUESTION}, generate timeline for what the question concerns and answer each sub-question. Then return the final answer.
\end{tcolorbox}
The Review action serves as a verification step. After an initial answer is generated, the model is asked to re-examine the original context and identify support sentences to justify the answer. If evidence is lacking or inconsistent, the model is encouraged to reconsider and revise its answer, enhancing factual consistency and robustness. This step mimics human self-checking behavior and helps reduce hallucinated or unsupported answers.
\begin{tcolorbox}[colframe=brown,
        width=1\linewidth,
        arc=1mm, 
        auto outer arc,
        title={\small Review Template},
        breakable,]
        In the context of \underline{STORY} and the \underline{QUESTION}, after obtain the answer, given the support sentences in original story and check if the answer is correct. If yes, return the answer again. If not, think again and return the right answer.
\end{tcolorbox}

We use a unified LLM prompt to guide the planner's decision-making process. The prompt is as follows.
\begin{tcolorbox}[colframe=brown,
        width=1\linewidth,
        arc=1mm, 
        auto outer arc,
        title={\small AdapTime Template},
        breakable,]
       Let's think step by step. First, if the question is complex, break the question down into several simple sub-questions. Then generate timeline for what the question concerns and answer each sub-question if the story's timeline is unclear. Then return the final answer. After obtaining the answer, if you are not sure, look for support sentences in the original story and check if the answer is correct. If yes, return the answer again. If not, think again and return the right answer.
\end{tcolorbox}


\section{Prompt Variation Experiments}
To further verify the robustness of our proposed method, we conducted prompt variation experiments using the latest DeepSeek-V3.2-exp model on TempReason-L2 and TempReason-L3. We tested several semantically equivalent variations of the prompt, changing only the wording and structure while keeping the intended logic unchanged. Results in Table \ref{tab:variant} show that reasoning performance remains stable across these variants, confirming that our method is not sensitive to specific prompt phrasing, and that the adaptive reasoning mechanism is the key to its effectiveness.
\begin{table}[t]
\centering
\setlength\tabcolsep{9.pt}
\renewcommand{\arraystretch}{1}
\resizebox{1\linewidth}{!}{
\begin{tabular}{l|cc|cc}
\toprule
\multirow{2}{*}{\textbf{Method}} 

& \multicolumn{2}{c|}{\textbf{TempReason-L2}} & \multicolumn{2}{c}{\textbf{TempReason-L3}} \\
& EM(\%) & F1(\%) & EM (\%)& F1(\%) \\
\midrule
AdapTime-Original & 50.9	& 55.2&	 52.5	& 55.1 \\

AdapTime-Variant1      &  50.3	 &54.9	& 52.8	& 56.0

\\
AdapTime-Variant2  &  50.6	 &55.2	 &52.1	& 55.5\\

AdapTime-Variant3      &  50.2	 &54.6	 &53.3	& 56.5

 \\

\bottomrule
\end{tabular}}
\caption{Performance of different prompt variants on TempReason-L2 and TempReason-L3.}
\label{tab:variant}
\end{table}

\section{Case Study}\label{app:case}



Table~\ref{tab:case-study} and~\ref{tab:case-study-cot} present a representative example illustrating how our proposed framework, AdapTime, improves temporal reasoning compared to the traditional Chain-of-Thought (CoT) approach. The task involves identifying where Mikhail Lomonosov was educated in January 1736, given a long narrative containing multiple temporally-anchored events and institutions.

The traditional CoT method attempts to construct a linear timeline by extracting and interpreting key events. While it correctly identifies that Lomonosov had completed his studies at the Slavic Greek Latin Academy and briefly attended the Kyiv-Mohyla Academy in 1735, it struggles to pinpoint his exact status in January 1736. Due to ambiguity around the transition period before his scholarship to the St. Petersburg Academy, the model ultimately predicts "Unknown", failing to commit to a specific answer.

In contrast, AdapTime applies a structured multi-step reasoning strategy:
Reformulate: It decomposes the original question into focused sub-questions, clarifying pre- and post-January 1736 educational phases.
Timeline Construction: It builds an explicit timeline of Lomonosov’s academic journey, aligning events with their corresponding years.
Answer Selection: Based on the timeline, AdapTime identifies that by January 1736, Lomonosov had already completed his studies in Moscow and was transitioning to the Academic University at the St. Petersburg Academy of Sciences.
Verification: It cross-verifies the answer with textual evidence, confirming that the scholarship was already granted in 1736 and that he was affiliated with the St. Petersburg Academy at that time.
This structured process enables AdapTime to resolve temporal ambiguities and ground its reasoning in anchored timelines, leading to a correct and explainable answer, while the traditional CoT remains uncertain. This case demonstrates how AdapTime significantly enhances the model's ability to reason over complex temporal contexts, especially when precise date alignment is required.

\begin{table*}[t]
\centering
\renewcommand{\arraystretch}{1.3}
\setlength{\tabcolsep}{10pt}
\begin{tabular}{p{4cm} | p{10cm}}
\hline
\textbf{Story} &
\textit{Mikhail Lomonosov was a Russian polymath, scientist and writer, who made important contributions to literature, education, and science. Among his discoveries were the atmosphere of Venus and the law of conservation of mass in chemical reactions. … In five years Lomonosov completed a twelve-year study course and in 1736, among 12 best graduates, was awarded a scholarship at the St. Petersburg Academy. He plunged into his studies and was rewarded with a four-year grant to study abroad, in Germany, first at the University of Marburg and then in Freiberg.The University of Marburg was among Europe's most important universities in the mid-18th century due to the presence of the philosopher Christian Wolff, a prominent figure of the German Enlightenment. Lomonosov became one of Wolff's students while at Marburg from November 1736 to July 1739. Both philosophically and as a science administrator, this connection would be the most influential of Lomonosov's life. …} \\

\textbf{Question} &
\textit{Where was Mikhail Vassilyevich Lomonosov educated in Jan, 1736?} \\

\textbf{Options} &
``Slavic Greek Latin Academy",
    ``Kyiv-Mohyla Academy",
    ``Kyiv-Mohyla Academy",
    ``Academic University at the St. Petersburg Academy of Sciences",
    ``Slavic Greek Latin Academy",
   ``Unknown" \\

\hline

\textbf{Step 1: Reformulate}
&Sub-question 1: Where was Lomonosov educated before January 1736?     - Lomonosov studied at the Slavic Greek Latin Academy in Moscow from 1730 to 1735.    - In 1735, he was sent to the Kyiv-Mohyla Academy for a short period but quickly returned to Moscow.  2. Sub-question 2: Where was Lomonosov in January 1736?    - In 1736, Lomonosov was among the 12 best graduates of the Slavic Greek Latin Academy and was awarded a scholarship at the St. Petersburg Academy.    - By January 1736, he had already completed his studies in Moscow and was preparing to move to St. Petersburg.
\\
\textbf{Step 2: Timeline}&
Timeline:     - 1730–1735: Slavic Greek Latin Academy (Moscow).    - 1735 (briefly): Kyiv-Mohyla Academy (Kiev).   - 1736: Awarded scholarship to the Academic University at the St. Petersburg Academy of Sciences.
\\
\textbf{Step 3: Answer}&
 Academic University at the St. Petersburg Academy of Sciences.
\\
\textbf{Step 4: Verification}&
The text states: In five years Lomonosov completed a twelve-year study course and in 1736, among 12 best graduates, was awarded a scholarship at the St. Petersburg Academy.    - This confirms that by 1736, he was no longer at the Kyiv-Mohyla Academy or the Slavic Greek Latin Academy but was affiliated with the St. Petersburg Academy. Thus, the correct answer is ( Academic University at the St. Petersburg Academy of Sciences ).
\\

\hline
\end{tabular}
\caption{Case study demonstrating temporal reasoning using the AdapTime framework.}
\label{tab:case-study}
\end{table*}
\begin{table*}[t]
\centering
\renewcommand{\arraystretch}{1.3}
\setlength{\tabcolsep}{10pt}
\begin{tabular}{p{3cm} | p{10cm}}
\hline

\textbf{Traditional CoT}& To determine where Mikhail Lomonosov was educated in January 1736, let's analyze the timeline of his education:1. 1724–1730: Lomonosov lived in his village, Mishaninskaya, and pursued basic education locally.  2.1730: He left for Moscow and was admitted to the **Slavic Greek Latin Academy**, where he studied for several years. 3. 1735: After excelling at the Slavic Greek Latin Academy, he was sent to the Kyiv-Mohyla Academy for a short period but quickly returned to Moscow due to dissatisfaction.  4. 1736: By this time, Lomonosov had completed his studies at the Slavic Greek Latin Academy and was among the top graduates. In 1736, he was awarded a scholarship to the Academic University at the St. Petersburg Academy of Sciences, marking the next phase of his education.  Since the question specifies January 1736, this falls after his time at the Slavic Greek Latin Academy and Kyiv-Mohyla Academy but before his enrollment at the Academic University in St. Petersburg. However, the text does not explicitly mention his whereabouts in January 1736, though it is likely he was still in Moscow transitioning to St. Petersburg.  Given the options:  - Slavic Greek Latin Academy: He had already completed his studies there by 1736.  Kyiv-Mohyla Academy: He attended this briefly in 1735 but had returned to Moscow by 1736.  Academic University at the St. Petersburg Academy of Sciences: His enrollment here began later in 1736, not January.  Thus, the most accurate answer is ( Unknown ).\\
\hline

\end{tabular}
\caption{Case study demonstrating temporal reasoning using the traditional CoT.}
\label{tab:case-study-cot}
\end{table*}

\end{document}